\documentclass[conference]{IEEEtran}
\ifCLASSINFOpdf
  \usepackage[pdftex]{graphicx}
\else
\fi
\usepackage[cmex10]{amsmath}
\usepackage{algorithmic}

\usepackage[numbers]{natbib}
\usepackage{notoccite}
\usepackage{amssymb}
\usepackage{color}
\usepackage{amsmath}
\usepackage{fancyvrb}
\usepackage{moresize}

\begin{document}
%
\title{\huge Text-mining the \textit{NeuroSynth} corpus using
Deep Boltzmann Machines
\vspace{-.85cm}}
\author{\IEEEauthorblockN{Ricardo Monti\IEEEauthorrefmark{1},
Romy Lorenz\IEEEauthorrefmark{2},
Robert Leech\IEEEauthorrefmark{2},
Christoforos Anagnostopoulos\IEEEauthorrefmark{1} and
Giovanni Montana\IEEEauthorrefmark{1}\IEEEauthorrefmark{3}}
\IEEEauthorblockA{\IEEEauthorrefmark{1}Department of Mathematics, Imperial College London}
\IEEEauthorblockA{\IEEEauthorrefmark{2}Computational, Cognitive and Clinical Neuroimaging Laboratory, Imperial College
London}
\IEEEauthorblockA{\IEEEauthorrefmark{3}Department of Biomedical Engineering, King’s College London
}}
\maketitle

\begin{abstract}
Large-scale automated 
meta-analysis of neuroimaging data has recently established
itself as an important tool in 
advancing our understanding of human brain function.
This research has been pioneered by \textit{NeuroSynth},
a database collecting both brain activation coordinates and 
associated text across a large cohort of 
neuroimaging research papers. 
One of the fundamental aspects of such meta-analysis
is text-mining. 
To date, word counts and more sophisticated methods such as 
Latent Dirichlet Allocation
have been proposed. 
In this work we present an unsupervised study of
the NeuroSynth text corpus using Deep Boltzmann Machines (DBMs). 
The use of DBMs yields several advantages over the aforementioned methods, 
principal among which is the fact that it yields both word and document 
embeddings in a high-dimensional vector space. 
Such embeddings serve to facilitate the use of traditional machine learning techniques 
on the text corpus. 
The proposed DBM model is shown to learn 
embeddings with a clear semantic structure. 

\end{abstract}



%
\IEEEpeerreviewmaketitle

\section{Introduction}

The study of the human brain using functional magnetic resonance imaging (fMRI)
has advanced rapidly in the last decades.
This has  provided significant 
insights into the relationship between architecture and function of
the human brain.
This is reflected in the number of published studies, which
has grown exponentially during this time. 
Consequently, a major challenge 
for the scientific community 
involves the efficient integration and
analysis of knowledge across this wide corpus of studies \citep{poldrack2012discovering}. 
This challenge has inspired attempts to 
automatically 
aggregate and analyze knowledge across the field of fMRI. 
In particular, NeuroSynth \citep{yarkoni2011large}
is a meta-analysis database collecting both brain activation coordinates and
the corresponding text across a range of over ten thousand studies. 
This has important 
applications in the analysis and interpretation of
fMRI data such as facilitating quantitative 
reverse inference \citep{poldrack2006can}.

The automated extraction of 
information from a collection of published neuroimaging studies 
is based on two fundamental pillars; 
the first of which involves generating
detailed statistical maps. In this 
work we focus on the second pillar; the 
extraction and analysis of 
semantic topics from text \citep{poldrack2012discovering}. 
Such methods look to employ text-mining methodologies to 
discover latent topics in 
the brain imaging literature. 
Such approaches can subsequently be combined with activation 
coordinates to examine the underlying mapping
between cognitive and neural states.


Recent attempts to model the semantic structure of the neuroimaging literature 
have focused on the use of Latent Dirichlet Allocation (LDA) models \citep{poldrack2012discovering}.
Such an approach is able to learn a pre-specified number 
of latent ``topics'' which generated observed text. 
In this work we present a related approach based on the 
use of Deep Boltzmann machines (DBMs). 
The motivation behind the use of DBMs over alternative 
text-mining approaches such as LDA is two-fold.
First, 
the use of restricted Boltzmann machines (RBMs), which 
are a special case of DBMs, has recently been shown to outperform 
LDA in terms of generalization performance \citep{hinton2009replicated}. This is hypothesized to be the result
of RBMs learning useful internal representations of the text corpus \citep{bengio2013representation}.
The presence of additional hidden layers in DBMs would serve to further facilitate the 
learning of internal representations. 
The second advantage of using DBMs is that such models yield an embedding of 
words or documents in a high-dimensional vector space.
Such embeddings 
are a crucial component of  
modern  natural language processing systems \citep{bengio2013representation} as they can 
be easily incorporated into traditional machine learning pipelines. 
Furthermore, the use of word embeddings can be employed to learn joint models across both text and the associated 
activation coordinates which is the ultimate objective of meta-analysis studies \citep{yarkoni2011large}. 

In this work we demonstrate that DBMs can be effectively employed to learn
the distribution of the NeuroSynth text corpus. Further,
the proposed model is able to learn embeddings of both individual words as 
well as entire documents. As motivation, Table \ref{table_wordEmbed} shows some of the clusters
obtained when $k$-means clustering is applied to word embeddings obtained
from the DBM model. The clusters display clear semantic context. 


\section{Materials and Methods}
\label{sec--MM}


\subsection{Deep topic models}
\label{sec--DBMs}
In this section we outline the models employed in this work.
We begin by introducing Restricted Boltzmann machines (RBMs), which 
serve as the building blocks of the deeper architectures considered in this work.
Extensions of RBMs to 
directly model word counts are discussed
before considering Deep Boltzmann machines (DBMs).

\subsubsection{Restricted Boltzmann machines}
RBMs are a class of undirected graphical models which specify a 
probability distribution over observed binary variables $v \in \{0,1\}^D$
and binary hidden variables $h \in \{0,1\}^F$. 
Formally, RBMs are energy based models which have a bipartite graph 
structure across visible and hidden variables. This structure is imposed in order to facilitate the 
learning of the models parameters which we discuss below. 

The following energy function is defined on any configuration of visible and hidden units:
\begin{equation}
 E(v,h; \theta) = -v^T W h - a^Tv -b^T h
\end{equation}
where $\theta = \{W, a, b\}$ are the parameters of the RBM which we wish to estimate. 
The probability of any given configuration $(v,h)$ is 
subsequently defined as $P(v,h; \theta) = \frac{1}{Z(\theta)} e^{-E(v,h; \theta)}$,
where $Z(\theta) = \sum_{v,h} e^{-E(v,h; \theta)}$ is normalizing constant.
Furthermore, the likelihood for any observation, $v$, can be obtained by
summing over binary hidden units:
\begin{equation}
\label{visible_likelihood}
 P(v; \theta) = \frac{1}{Z(\theta)} \sum_h e^{-E(v,h;\theta)}.
\end{equation}

Parameter learning in RBMs is typically achieved via
performing gradient descent on the log-likelihood 
over observed data. 
From equation (\ref{visible_likelihood}),
the training data log-likelihood is composed of  
a \textit{positive} term, $\phi^+ = \mbox{log} \sum_h e^{-E(v,h; \theta)}$,
and a \textit{negative} term, $\phi^- = \mbox{log}~ Z(\theta)$ \citep{tieleman2008training}.
The derivate with respect to the 
positive term corresponds to an expectation over the data dependent distribution of hidden variables, 
which can be easily computed due to the bipartite structure of RBMs.
However, the derivative of the negative term involves an 
expectation over the distribution of both visible and hidden units under the proposed model which is intractable. 
This expectation is typically approximated by looking to sample from this distribution using MCMC.
Starting with visible units, Gibbs sampling is applied $k$ times in order to
obtain an unbiased sample of the gradient in a procedure known as Contrastive Divergence \citep{hinton2002training}.
Letting $k \rightarrow \infty$ recovers  maximum likelihood, however in practice 
it has been shown empirically that setting $k=1$ performs well.

\begin{table}[t]
    \caption{ Examples of semantic clusters learnt by deep topic model }
    \begin{tabular}{ |  p{8.4cm} |}
    \hline
    \textbf{Associated vocabulary} \\ \hline
     memory, retrieval, encoding, 
    hippocampus, hippocampal, episodic, items, recall, memories, recollection, item, familiarity, autobiographical\\ \hline 
     language, semantic, words, speech, word, reading, verbal, phonological, lexical, linguistic, naming, fluency, verbs, english \\ \hline
     adults, age, children, years, older, young, development, adolescents, developmental, aging, sleep, adult, late, younger, blind, childhood, hearing, adolescence \\ \hline 
     emotional, amygdala, social, negative, faces, face, emotion, neutral, affective, facial, anxiety, fear, expressions, regulation, emotions, ofc, valence, personality, arousal, fearful, trait, threat, sad, happy, mood, empathy, moral, person, traits, communication \\ \hline
     patients, controls, schizophrenia, disorder, deficits, disease, abnormalities, symptoms, impaired, impairment, adhd, alterations, dysfunction, mdd, abnormal, atrophy, patient, ptsd, severity, mci, damage, bipolar, lesions, impairments, deficit, depressive, ocd, mild, syndrome, symptom, elderly, dementia, epilepsy, poor, pathophysiology \\ \hline
    \end{tabular}
    \label{table_wordEmbed}
    \end{table}

\subsubsection{Replicated Softmax model}

The aforementioned RBM model can be employed 
when the objective is to learn the  probability 
over binary visible variables. 
In the context of modeling documents it is possible to 
treat the occurrence  of words at specific locations in the text as binary variables.
In this case the observations correspond  to a binary incidence matrix $V \in \{ 0,1\}^{N\times D}$
where $V_{n, d}$=1 if the $n$th word in the document takes the $d$th value. 

While such an approach is able to model the order of words, there
is an explosion in the number of parameters. 
The replicated softmax RBM takes a more parsimonious alternative, directly 
modeling the word counts, $\hat v_d = \sum_n V_{n,d}$ \citep{hinton2009replicated}. 
In such a setting visible units $\hat v \in \mathbb{N}^D$
correspond to a vector of words counts for each document.
Note that 
$D$ corresponds to the size 
of the vocabulary. 

The energy of a state $(\hat v,h)$ is defined as:
\begin{equation}
 E(\hat v,h; \theta) = - \hat v^T W h - a^T \hat v -M \cdot b^T h,
\end{equation}
where $M = \sum_{d} \hat v_d$ is the total number of words in a document. 
As with a standard RBM, learning proceeds via Contrastive Divergence.
Such models can be interpreted as learning a distribution over word histograms of documents.

%
%

\subsubsection{Deep Boltzmann machines}
DBMs are 
extensions of RBMs to allow for multiple layers of hidden variables. 
Such models have the capability of learning 
internal representations of the data which are increasing complex \citep{salakhutdinov2009deep}. 
Throughout this work we consider a two-layer DBM
with multinomial visible variables and binary hidden variables.
Such a model is associated with the following energy function\footnote{we have excluded bias terms for clarity}:
\begin{equation}
 E(\hat v, h^1, h^2) = -\hat v^T W^1 h^1 - {h^1}^T W^2 h^2
\end{equation}
where we write $h^1$ and $h^2$ to denote the first and second layer of binary hidden variables respectively.
Similarly, parameters $\theta = \{W^1, W^2\}$ represent the 
symmetric interaction terms between
visible-to-hidden and hidden-to-hidden variables.
Analogous to equation (\ref{visible_likelihood}), the probability assigned to a visible 
vector, $\hat v$ is defined as:
\begin{equation}
\label{visible_likelihood_DBM}
 P(\hat v; \theta) = \frac{1}{Z(\theta)} \sum_{h^1, h^2} e^{-E(\hat v,h^1, h^2;\theta)}
\end{equation}
Furthermore, due to the bipartite across layers the conditional distributions of
each of the layers can be computed in closed form. This allows for the use of 
persistent Markov Chains \citep{tieleman2008training} to estimate the intractable model expectations.  
Naive mean-field variational inference is then used to approximate the data-dependent expectations. 
For further details we refer readers to \cite{salakhutdinov2009deep}. 

In practice, appropriate initialization of parameters is crucial to 
the success of deep models. 
\cite{salakhutdinov2009deep} propose a greedy, layer-by-layer 
pretraining algorithm for DBMs. This involves iteratively stacking RBMs,
with the small caveat that bottom-up (likewise top-down) contributions 
from the bottom (top) layer  should be doubled during pretraining.


\subsubsection{Model selection}
\label{sec--AIS}
Selecting the number of hidden units within each layer of a DBM is a 
non-trivial task. 
The difficulty of such an approach arises from the need to estimate 
the (typically intractable) partition function $Z(\theta)$ for the entire model. 
As $Z(\theta)$ depends on both the parameters as well as number of hidden units, 
it must be calculated in order to perform model comparison. 

Importance sampling is often employed to estimate properties 
of distributions known only up to a normalizing constant 
using samples from a known distribution.
However, for importance sampling to yield a reliable estimate 
the known proposal distribution must resemble the target distribution.
In the context of high-dimensional RBMs finding such a proposal distribution is challenging.
In order to address this challenge, \cite{salakhutdinov2008quantitative} propose the 
use of annealed importance sampling (AIS). Here a sequence of auxiliary 
proposal distributions are 
defined which iteratively approximate the target distribution. 

Due to the bipartite structure of RBMs,
it is easy to transition across the intermediate distributions (in practice we apply one iteration of Gibbs sampling).
In this fashion it is possible to begin with a sample from a uniform RBM (with partition $Z_0 = 2^F$),
which we propagate through auxiliary distributions \citep{hinton2009replicated}. 

In this work a greedy, layer-by-layer approach was taken to select the model architecture. 
As a result, the bottom layer RBM was trained using a range of hidden units. The architecture 
which yielded the maximum likelihood across a 
held-out validation set was selected. 
The hidden activation from this RBM was subsequently provided as input for the top layer RBM 
and the process was repeated.

\subsection{Dataset}
\label{sec--Dataset}

The NeuroSynth text corpus was employed in this work.
While the original corpus 
contains word frequencies over the entire text for each publication, 
in this work only the publication abstracts were employed. This served to reduce the range of vocabulary employed 
and was motivated by our belief that much of the 
semantic structure present in a publication would also be present in the corresponding abstract. 
Abstracts were collected for 10574 publications using the PUBMED API resulting in a mean document length of
80 words ($\pm 25$ words).  
%
Standard preprocessing was applied to the text corpus.  Stop words were removed,
as well as words which did not occur with sufficient frequency (fewer than 50 occurrences throughout
the corpus). This resulted in a  
vocabulary of approximately two thousand words, of which the 1000 words which occurred most 
frequently were retained (corresponding to over $80\%$ of terms). 
The dataset was split into a
training set consisting of 9516 documents and a test set with the remaining
1058 documents. 

\section{Results}

\subsection{Model architecture and implementation details}
A two-layer DBM was employed consisting of a visible layer of 
multinomial visible units followed by two binary hidden layers with 
50 units each.
During pretraining and model selection 
RBMs where trained using CD$-1$. 
In addition,
dropout was employed as a form of regularization 
with hidden units retained with probability $0.9$. 

The architecture was selected 
by minimizing the negative log-likelihood 
over a held out validation dataset in a greedy manner as described previously. 
Briefly, AIS was employed to 
estimate the partition function for each RBM. Five thousand auxiliary distributions 
were employed (specified by 
uniformly spaced inverse temperatures) and
estimates were averaged over five hundred runs.
Finally, the DBM was initialized to weights learnt during pretraining and
trained as described in \citep{salakhutdinov2009deep}.



The proposed DBM model can be used to obtain 
both word as well as document embeddings in a high-dimensional vector space\footnote{in our case 
the embedding will be in $\mathbb{R}^{50}$ as the top layer has 50 hidden units}.
In the remainder of this section we study both the word and document embeddings obtained 
from the proposed DBM model.

\subsection{Word embeddings}
The proposed DBM model can be employed to obtain a 
high-dimensional embeddings for each word in our vocabulary.
The word clusters shown in Table \ref{table_wordEmbed}
can then be obtained by 
by applying $k$-means clustering to the word embeddings.
The number of clusters was selected based on silhouette scores. 

Further, Figure [\ref{BigFig}i] shows a 2D visualization of word embeddings using t-SNE \citep{van2008visualizing}.
Three sections of the embedding have been highlighted. 
Regions A and C showcase embeddings for terms relating to emotion and memory respectively.
It is important to note that the relevant brain regions are contained in 
this sections (i.e., the amygdala and orbitofrontal cortex in region A and 
the hippocampus in region C).
Meanwhile region B contains terms relating to age and development. 

Finally, an alternative manner of demonstrating the DBM model has obtained a good
estimate of the probability distribution is to consider one-step reconstructions.
Some examples are provided in Table \ref{table_recon}. The input words where 
employed to obtain a distribution over hidden units at the top level. This distribution was then
employed to obtain a distribution over words. The words with highest probability mass
are shown in the right column.

\begin{table}[t]
    \caption{Examples of one-step reconstruction}
    \begin{tabular}{ | l | p{7cm} |}
    \hline
    \textbf{Input} & \textbf{One-step reconstruction} \\ \hline
    memory & memory, working, recall, performance, retrieval, verbal, load, semantic, recognition, task \\
    \hline
    emotion & social, emotion, emotional, regions, ofc, brain, affective, gray, traits, amygdala \\ \hline 
    face & social, facial, faces, face, emotional, processing, regions, functional, brain, cortex \\ \hline
    disorder & patients, mdd, disorder, adhd, abnormalities, controls, brain, matter, alterations, structural \\ \hline
    mode & network, default, connectivity, brain, regions, cognitive, functional, mode, activity, cortex \\ \hline
    \end{tabular}
    \label{table_recon}

    \end{table}

\subsection{Document embeddings}
Document embeddings are obtained in analogous fashion by providing the entire document word vector as input to the DBM. 
By clustering document embeddings and leveraging the activation maps within the NeuroSynth
database, we are able to study the activations associated with each cluster.

\begin{figure*}[!t]
    \centering
    \includegraphics[width=\textwidth]{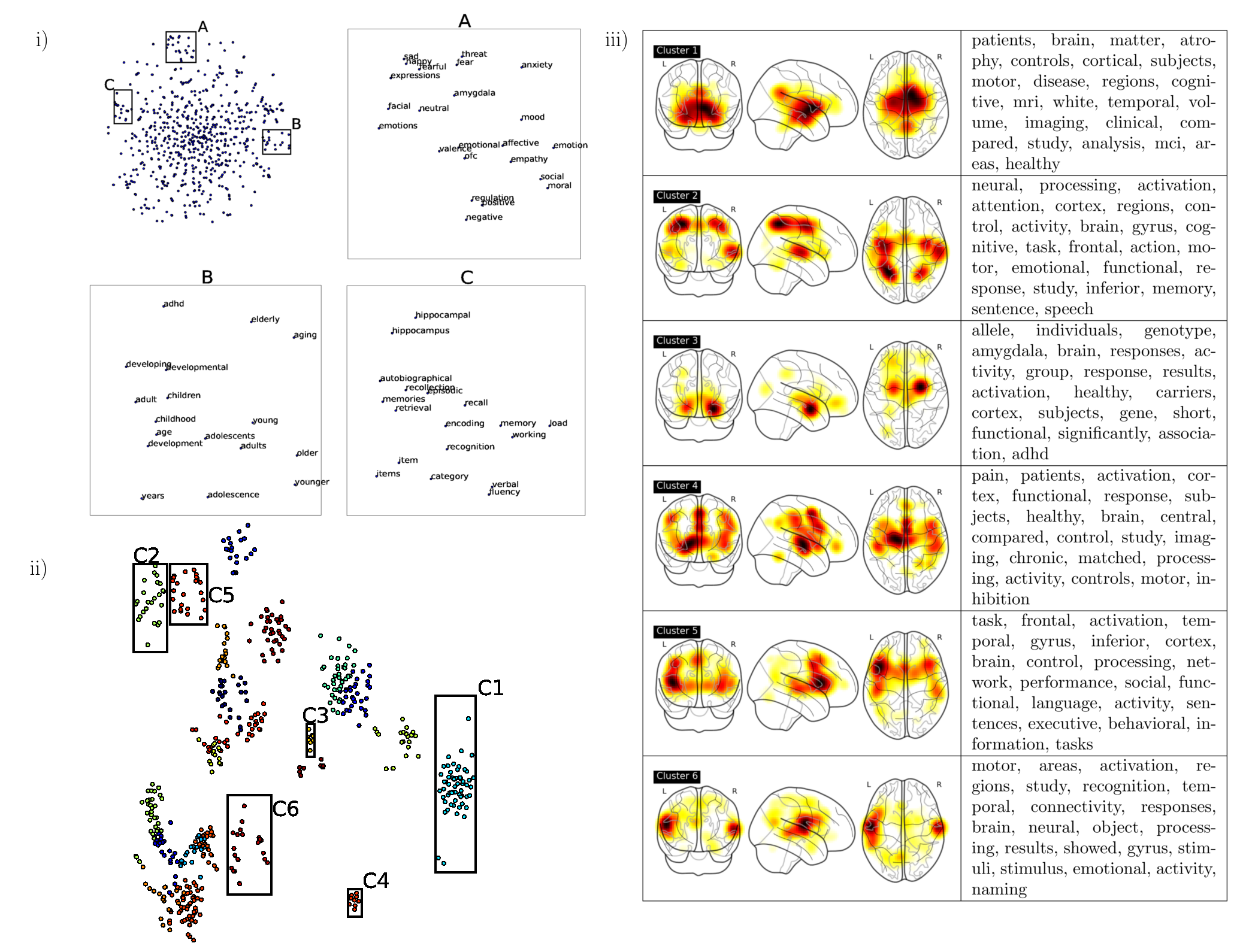}
    \caption{\textbf{i)} The top left panel shows the result of applying t-SNE on 
    word embeddings obtained from the DBM model. Three regions have been highlighted 
    are are shown in greater detail in the remaining three panels. It can be seen that regions A and C correspond to 
    emotion and memory related terms respectively while region C contains terms associated with aging and
    development. \textbf{ii)} A subset of the two dimensional embedding obtained from applying t-SNE on 
    document embedding. \textbf{iii)} Activation maps (left column) are shown for several of the highlighted 
    clusters shown in ii) together the with the most frequently occurring terms (right column). 
  }
    \label{BigFig}
\end{figure*}

Figure [\ref{BigFig}ii] shows a subset of the 2D embeddings obtained using t-SNE over all document embeddings.
As before, $k$-means 
clustering was employed to cluster documents according to their 
associated high-dimensional representations (again silhouette scores used to select number of clusters). 
It is then possible to study the
reported functional activations of all publications within a given cluster. 
Following \cite{poldrack2012discovering}, this was achieved by 
convolving all activations within a cluster (i.e., all activations reported by documents in a cluster)
with a 10mm Gaussian kernel. This resulted in a mean activation map for all documents within a given cluster.
The peak activations, together with the most frequently occurring words 
are shown for six clusters in Figure [\ref{BigFig}iii]\footnote{figure produced using nilearn module \citep{abraham2014machine}.}. 
The activation maps highlight key functional networks and regions, for example clusters four and six identify the 
pain and motor regions respectively. The clusters also appear to identify
pathologies. For example cluster one 
appears to be related to 
cognitive impairment and atrophy. 
Furthermore, 
it is interesting to note that 
spatially adjacent clusters share some similarities.
For example, clusters two and five both show frontal activation.

\section{Discussion and future work}

In this paper we have demonstrated the use of DBMs in modeling a text corpus 
composed of abstracts from neuroscientific publications. 
The proposed DBM model is able to yield a vector representation of both 
individual words as well as entire documents. 
Such representations are advantageous for many reasons, for example they can be employed 
to cluster the words or documents.

Further, by combining the abstracts with the NeuroSynth corpus, we are able to 
study whether the activation maps associated with each cluster. 
While only exploratory results are presented in this work, 
future work will look simultaneously model both text and activations, 
thereby facilitating formal inference. 
A further exciting application would be to leverage 
document embeddings to
inform novel machine learning applications in neuroscience, such as the 
recently proposed \textit{Automated Neuroscientist} \citep{lorenz2016automatic}.

\newpage 
{\ssmall 
\bibliographystyle{unsrtnat}
\bibliography{ref}
}

\end{document}